\documentclass[conference]{IEEEtran}
\IEEEoverridecommandlockouts
\usepackage{cite}
\usepackage{amsmath,amssymb,amsfonts}
\usepackage{algorithmic}
\usepackage{graphicx}
\usepackage{textcomp}
\usepackage{xcolor}

\usepackage{lscape}
\usepackage{multirow}

\def\BibTeX{{\rm B\kern-.05em{\sc i\kern-.025em b}\kern-.08em
    T\kern-.1667em\lower.7ex\hbox{E}\kern-.125emX}}
\begin{document}

\title{Low Precision Quantization-aware Training in Spiking Neural Networks with Differentiable Quantization Function}


 \author{\IEEEauthorblockN{Ayan Shymyrbay$^1$, Mohammed E. Fouda$^2$, and Ahmed Eltawil$^1$}
 \IEEEauthorblockA{$^1$ Department of ECE, CEMSE Division, King Abdullah University of Science and Technology, Saudi Arabia\\
 $^2$Center for Embedded \& Cyber-physical Systems, University of California-Irvine, Irvine, CA, USA 92697-2625. }}

\maketitle

\begin{abstract}
Deep neural networks have been proven to be highly effective tools in various domains, yet their computational and memory costs restrict them from being widely deployed on portable devices. The recent rapid increase of edge computing devices has led to an active search for techniques to address the above-mentioned limitations of machine learning frameworks. The quantization of artificial neural networks (ANNs), which converts the full-precision synaptic weights into low-bit versions, emerged as one of the solutions. At the same time, spiking neural networks (SNNs) have become an attractive alternative to conventional ANNs due to their temporal information processing capability, energy efficiency, and high biological plausibility. Despite being driven by the same motivation, the simultaneous utilization of both concepts has yet to be thoroughly studied. Therefore, this work aims to bridge the gap between recent progress in quantized neural networks and SNNs. It presents an extensive study on the performance of the quantization function, represented as a linear combination of sigmoid functions, exploited in low-bit weight quantization in SNNs. The presented quantization function demonstrates the state-of-the-art performance on four popular benchmarks, CIFAR10-DVS, DVS128 Gesture, N-Caltech101, and N-MNIST, for binary networks (64.05\%, 95.45\%, 68.71\%, and 99.43\% respectively) with small accuracy drops and up to 31$\times$ memory savings, which outperforms existing methods.
\end{abstract}


\section{Introduction}
\label{sec:Introduction}
Spiking neural networks (SNNs) have recently become a popular research field for machine learning enthusiasts and neuromorphic hardware engineers due to their temporal data processing, biological plausibility, and energy efficiency \cite{31Bouvier2019, 51Fang2021, Fouda}. These properties make them ideal for reducing hardware implementation costs on resource-limited devices for real-world artificial intelligence (AI) applications \cite{2Pfeiffer2018}. Furthermore, recent advancements in event-based audio and vision sensors have opened up many opportunities for various domain based applications\cite{2Pfeiffer2018, Ibrahim}. Choosing the right neural network (NN) type can significantly impact resource utilization, but there are additional ways to alleviate the disadvantages of deploying large models on edge devices. While increasing the model's size leads to better performance, it also results in higher memory utilization and inference time. There are three widely used NN compression methods used to reduce the memory and computation costs of the model without significant performance degradation: quantization, pruning, and knowledge distillation\cite{15Menghani2021, Guo1, Guo2}. 

Quantization is a process of reducing the number of bits to represent synaptic weights. Storing and operating on reduced bit precision weights allows for significantly improved memory savings and power efficiency \cite{15Menghani2021}. Quantization methods are divided into two types depending on whether the quantization process occurs during or after network training. These types are quantization-aware training (QAT) and post-training quantization (PTQ). QAT usually results in lower accuracy loss than PTQ. However, training a quantized model from scratch increases the quantization time \cite{26Weng2021}. In PTQ, one can choose any pre-trained model and perform quantization much faster; however, getting good performance in low-bit precision is difficult \cite{26Weng2021}. At the same time, doing QAT can reduce the bit precision down to 1 bit. There are different scenarios where each of these methods can be preferable. QAT is preferable when the quantized model is to be deployed for a long time, and hardware efficiency and accuracy are the main goals. PTQ is a better option when there is insufficient data to train the model and if fast and simple quantization is needed. The state-of-the-art has used various techniques to quantize SNN parameters in either PTQ or QAT manner.

While weight quantization has been applied to different artificial neural networks (ANNs), its benefits on SNNs are yet to be thoroughly studied. Limited works have reported applying weight quantization in SNNs. Amir et al.\cite{49Amir2017} use a deterministic rounding function to quantize convolutional SNN to be deployed on a TrueNorth neuromorphic chip. Eshraghian and Lu\cite{48Eshraghian2022} explore how adjusting the firing threshold of SNN can help with the deterministic binarization of the weights. Schaefer and Joshi\cite{27Schaefer2020} train SNN models with integer weights and other parameters having variable precision using a deterministic rounding function. Lui and Neftci\cite{41Lui2021} show how a layer-wise Hessian trace analysis can quantify how a change in weights influences the loss. The authors claim that this metric can help with the intelligent allocation of a layer-specific bit-precision while training SNNs. Rathi et al.\cite{43Rathi2019} perform quantization and pruning simultaneously by exploiting the natural advantages of the spike timing-dependent plasticity (STDP) learning rule. The authors use weight distribution to create weight groups and quantize them by averaging. Putra and Shafique\cite{40Putra2021} propose a framework for quantizing SNN parameters in PTQ and QAT using truncation, round to the nearest, and stochastic rounding techniques. Hu et al.\cite{45Hu2021} present an STDP-based weight quantization technique that uses a round function. Kheradpisheh et al.\cite{46Kheradpisheh2022} use full-precision parameters in the backward pass and signs of the synaptic weights in the forward pass to binarize synaptic weights. Another binarization method is proposed by Kim et al.\cite{47Kim2021} in which binary weights are learned in an event-based manner with the help of a constraint function and a Lagrange multiplier. The authors also use event-driven random backpropagation instead of STDP. 

Most existing quantization methods are based on non-differentiable quantization functions, making it infeasible to compute gradients to train NNs by the most widely used training algorithm - the gradient descent algorithm \cite{29Qin2020}. The quantization functions are approximated for gradient calculation at the backward propagation to make the quantization methods compatible with the gradient descent algorithm. This approximation function is called straight-through estimator (STE) \cite{30Hassan2020, 29Qin2020}. However, since the approximation function cannot fully describe the quantization function, it produces different gradients from the $true$ gradients. SNNs also operate on discrete, non-differentiable spiking signals. Thus, the quantization of SNN models involves two non-differentiable functions: spikes and quantization. The non-differentiable spike function is usually solved through the surrogate gradient method (SGM), which approximates the spikes with differentiable surrogate analog in the backward pass to train the network \cite{37Neftci2019}. As both non-differentiable functions are approximated by SGM and STE for the training, quantized SNN models may encounter significant gradient mismatch and fail to achieve competitive performance compared to a full-precision counterpart. 

In this work, we propose adopting the QAT-based ANN differentiable quantization function by Yang et al. \cite{50Yang2019} for SNN quantization to reduce the number of gradient approximations involved in training and quantizing SNN models. This quantization function is based on a linear combination of sigmoid functions. Due to the benefits of using SNNs and quantization together, this paper aims to provide the findings on the performance of the mentioned quantization function for SNN quantization.

The main contributions of this paper are:

\begin{enumerate}
    \item We present the framework for quantizing SNN models using a differentiable quantization function based on a linear combination of sigmoid functions. The proposed QAT method demonstrates the effectiveness of using a differentiable quantization function to reduce the number of approximations in the training and quantization of SNN models. 
    
    \item We evaluate the quantization method on popular benchmark datasets such as CIFAR10-DVS, DVS128 Gesture, N-Caltech101, and N-MNIST. We find that the presented quantization method outperforms the state-of-the-art methods in terms of accuracy and memory savings.
\end{enumerate}

The paper is organized as follows: Section \ref{sec:Method} provides the mathematical modeling of spiking neurons and describes the quantization method, represented as a linear combination of sigmoid functions, as well as formulates it in the scope of SNN quantization, Section \ref{sec:Setup} provides the experimental setup, Section \ref{sec:Results} presents and discusses the obtained results, and Section \ref{sec:Conclusion} summarizes the findings on using the proposed differentiable quantization function for SNN quantization.

\section{Proposed Method}
\label{sec:Method}
\subsection{Spiking Neural Network}
In a real neuron, impulse transmission is determined by differential equations corresponding to the biophysical processes of potential formation on the neuron membrane.
A leaky integrate-and-fire (LIF) neuron is one of the most widely used mathematical models employed to accurately mimic the biological behavior of the neuron\cite{33Gerstner2014, 34Abusnaina2014, 51Fang2021}. This model assumes that the input, $X(t)$, comes as a voltage increment, then causes the hidden state (membrane potential), $H(t)$, to be updated, which in its turn causes the output, $S(t)$, to show spikes. $V(t)$ represents the membrane potential after the spike trigger. Its behavior can be described using the following equations:

\begin{equation}
\label{eq:charge}
   H(t)=f(V(t-1),X(t)) 
\end{equation}
\begin{equation}
\label{eq:discharge}
\begin{split}
    S(t) & = g(H(t)-V_{threshold}) \\
    & =\Theta(H(t)-V_{threshold})
\end{split}
\end{equation}
\begin{equation}
\label{eq:reset}
    V(t)=H(t)(1-S(t))+V_{reset} S(t)
\end{equation}

\noindent
where $X(t)$ is the input to the neuron at timestep $t$, $V_{threshold}$ is the threshold for spike firing, and $V_{reset}$ is the potential to which the neuron returns after the spike.
$f(V(t-1),X(t))$ denotes the state update equation of the neuron. For the LIF neuron, the following update function is used, where $\tau$ is a membrane time constant: 
\begin{equation}
\begin{split}
    H(t) & = f(V(t-1),X(t)) \\
    & =V(t-1)- \frac{1}{\tau} ((V(t-1)+V_{reset} )+X(t))
\end{split}
\end{equation}

For the spike generation function $g(t)$, the Heaviside step function $\Theta(x)$ is used, which has the following definition:

\begin{equation}
    \Theta(x) = \begin{cases}
1, & \text{for } x \geq 0 \\
0, & \text{for } x < 0  \\
\end{cases}
\end{equation}

\subsection{Quantization}
The quantization function presented in \cite{50Yang2019} is represented by a linear combination of sigmoid functions to eliminate gradient approximations while training and quantizing ANNs. This work uses this method to reduce the number of approximations for gradient computations while training and quantizing SNN models by the SGM rule. At inference, the quantization function is formulated as a combination of several unit-step functions with respective scaling and biasing:

\begin{equation}
\label{eq:quant_func}
    W_Q = \sum_{i=1}^{n} s_i \Theta (\beta W - b_i) - o
\end{equation}

In Eq. (\ref{eq:quant_func}), $W$ is the full-precision weight that needs to be quantized, and $W_Q$ is the discrete output value, $\Theta$ is the unit step function. The possible discrete values that $W_Q$ can take are pre-defined. $\beta$ is an input scale factor that maps the ranges of $W$ to the range of $W_Q$. The number of unit step functions is defined by the necessary number of quantization levels, $(n+1)$. $s_i$ represents the difference between adjacent quantization levels, and $b_i$ defines their border. The term $o$ is used to place the center of the quantization function at $0$,  $o = \frac{1}{2} \sum_{i=1}^{n} s_i$. Since the step function is not differentiable, training feedforward networks directly with the backpropagation method is not feasible. Step functions are replaced with sigmoid functions for training, while step functions are kept for inference. The following sigmoid function has a term $T$ (called ‘temperature’) that controls the gap between two quantization levels:

\begin{equation}
\label{eq: quant_func_2}
    \sigma (Tx) = \frac{1}{1 + e^{-Tx}}
\end{equation}

In Eq. (\ref{eq: quant_func_2}), increasing $T$ makes the border between quantization levels sharper, making this sigmoid function closer to a step function while remaining differentiable. However, choosing a large value for $T$ at the initial epochs of training yields poor training since most of the gradients will be zero. Hence, networks should be trained with small values of $T$ at the beginning, which is later increased by a small amount during each epoch. 

\subsection{Forward pass}
During forward propagation, weights in each layer of the network are mapped to the discrete integers according to the quantization function. In the inference stage, a step function, Eq. (\ref{eq: inference}), is used, and in the training stage, it is replaced by a sigmoid function, Eq. (\ref{eq: training}).

\begin{equation}
\label{eq: inference}
    W_Q = \alpha \left(\sum_{i=1}^{n} s_i \Theta (\beta W - b_i) - o\right)
\end{equation}

\begin{equation}
\label{eq: training}
W_Q = \alpha \left(\sum_{i=1}^{n} s_i \sigma (\beta W - b_i) - o\right)
\end{equation}

There are two kinds of parameters in Eq. (\ref{eq: training}): (1) learned during training and (2) pre-defined. Learned parameters are $\alpha$ and $\beta$, scale factors of output and input, respectively. Pre-defined parameters are the set of discrete integers representing quantization levels, and $b_i$, $T$, $s_i$, $o$. Each layer has different learned parameters and can have different pre-defined parameters. 

\subsection{Backward pass}
The gradients of loss $L$ should be backpropagated during the training stage, considering the quantization function. The mean squared error (MSE) loss is utilized as a loss function, which can be defined as:

\begin{equation}
    L = \frac{1}{T} \sum_{t=0}^{T-1} \cdot \frac{1}{N}\sum_{n=0}^{N-1} (S(t,n) - y(t,n))^2
\end{equation}

\noindent
where $T$ is timesteps, $N$ is a number of classes, $S(t,n)$ is a output spike and $y(t,n)$ is a true label at a timestep $t$ for class $n$. The weighted inputs from $l-1$ layer can be defined as $X^l(t) = W^{l-1}_Q I^l(t)$, where $W^{l-1}_Q$ is a quantized weight matrix and $I^l(t)$ is input spike matrix. Using Eq. (\ref{eq:charge}) - (\ref{eq:discharge}), we can formulate the gradients of loss as following:
\begin{equation}
\label{eq:loss1}
\begin{split}
    \frac{\partial L}{\partial W^{l-1}_Q} &= \sum_{t=0}^{T-1} \frac{\partial L}{\partial H^l(t)} \cdot \frac{\partial H^l(t)}{\partial W^l_Q} \\
    &=  \sum_{t=0}^{T-1} \frac{\partial L}{\partial S^l(t)} \cdot \frac{\partial S^l(t)}{\partial H^l(t)} \cdot \frac{\partial H^l(t)}{\partial X^l(t)} \cdot I^l(t)
\end{split}
\end{equation}

\noindent
where $ \frac{\partial S^l(t)}{\partial H^l(t)} = \Theta' (H^l(t) - V^l_{threshold})$ and $\frac{\partial H^l(t)}{\partial X^l(t)} = \frac{1}{\tau}$. Substituting these terms into Eq. (\ref{eq:loss1}) yields:

\begin{equation}
    \frac{\partial L}{\partial W^{l-1}_Q} =  \sum_{t=0}^{T-1} \frac{\partial L}{\partial S^l(t)} \cdot \Theta' (H^l(t) - V^l_{threshold}) \cdot \frac{1}{\tau} \cdot I^l(t)
\end{equation}

According to the SGM, $\Theta$ is replaced by a differentiable function for gradient calculation during the backward pass.

Three network parameters are learned during the training stage: synaptic weights, input scale factor, and output scale factor. The gradients in a $d$ layer w.r.t. these parameters and quantization function can be computed as shown below:

\begin{equation}
    \frac{\partial L}{\partial W^d} = \frac{\partial L}{\partial W^d_Q} \cdot \sum_{i=1}^{n} \frac{T\beta}{\alpha s_i} g_{d}^{i} (\alpha s_i -  g_{d}^{i})
\end{equation}

\begin{equation}
    \frac{\partial L}{\partial \alpha} = \sum_{d = 1}^{D} \frac{\partial L}{\partial W^d_Q} \cdot \frac{W^d_Q}{\alpha}
\end{equation}

\begin{equation}
    \frac{\partial L}{\partial \beta} = \sum_{d = 1}^{D} \frac{\partial L}{\partial W^d_Q} \cdot \sum_{i=1}^{n} \frac{T W^d_Q}{\alpha s_i} g_{d}^{i} (\alpha s_i -  g_{d}^{i})
\end{equation}

\noindent
where $g_{d}^{i} = \sigma(T(\beta W^d_Q - b_i))$. 

\section{Experimental Setup}
\label{sec:Setup}
We built a platform for SNN quantization using PyTorch. We utilized SpikingJelly \cite{38Fang2020} package for working with GPU-accelerated spiking neuron models. In this platform, one can set different quantization levels, choose the temperature rate, and control the simulation hyperparameters.

\begin{figure*}[t!]
\begin{center}
\includegraphics[width=14cm]{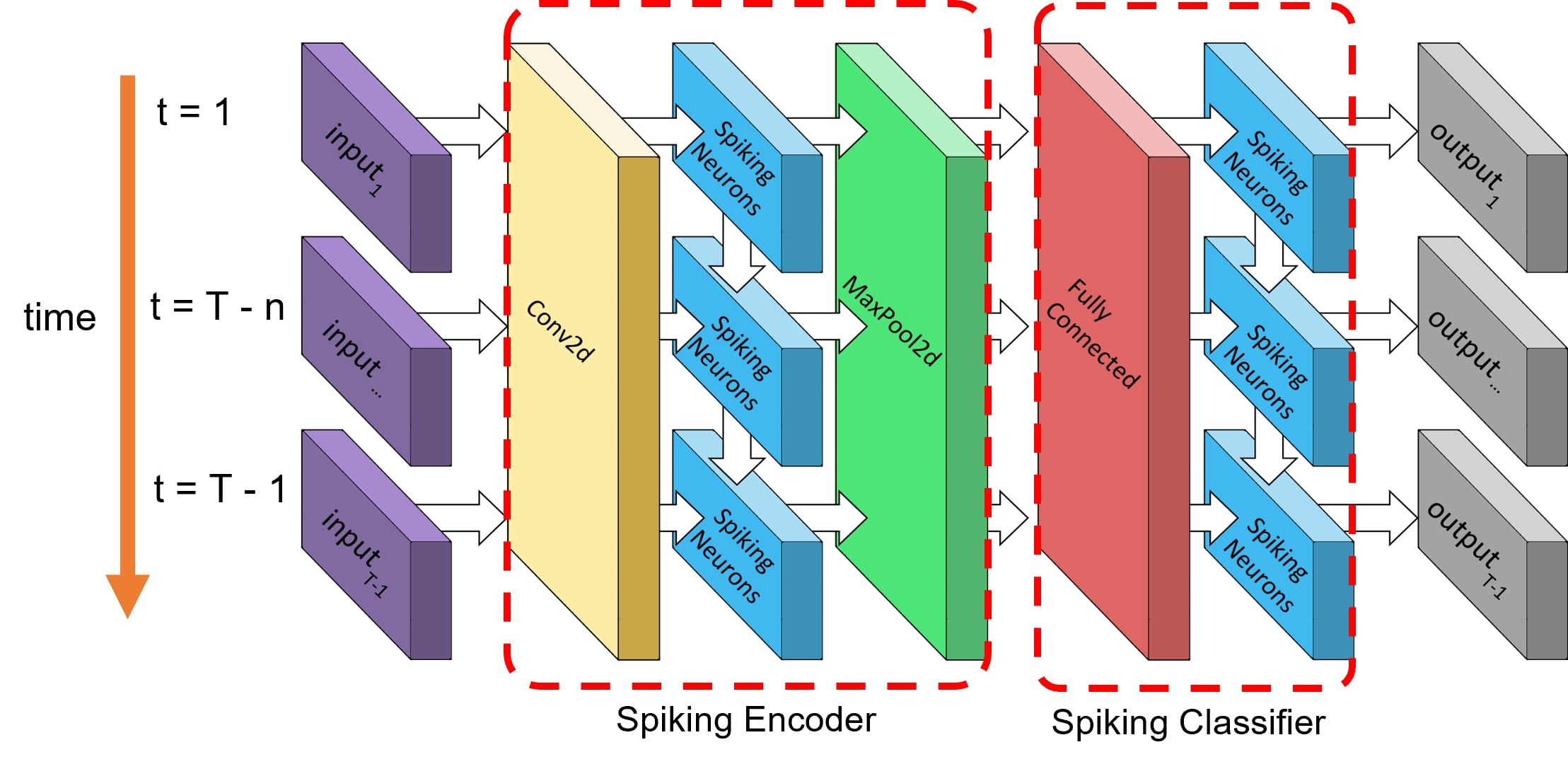} 
\end{center}
\caption{SNN model with a spiking encoder network and a classifier network.} \label{fig:Network}
\end{figure*}

\subsection{Datasets}
The performance of SNN models is evaluated on four widely used datasets: CIFAR10-DVS, DVS128 Gesture, N-Caltech101, and N-MNIST. The event-to-frame integrating method \cite{51Fang2021} is used to pre-process these datasets, which are initially represented in address event representation (AER) format. This method splits the event data into slices and integrates each into a single frame. The event data is sliced into ten frames for all datasets, representing $T=10$ timesteps.

\begin{itemize}
    \item CIFAR10-DVS dataset consists of 10,000 samples corresponding to 10 classes, with 1,000 samples in each class \cite{52Krizhevsky2022}. Since there are no direct train and test splits, this dataset is divided randomly into train and test sets in an 80/20 ratio, respectively.
    
    \item DVS128 Gesture dataset consists of 11 hand gestures collected from 29 subjects under three different illuminations \cite{49Amir2017}. The dataset comes with the train, and test splits in the ratio of 80/20 (1,176/288 samples).
    
    \item	N-Caltech101 dataset is a spiking version of Caltech101, consisting of 101 classes \cite{53Orchard2015}. In this work, the dataset is split randomly into train and test sets in the ratio of 80/20 (7,000/1,709 samples).
    
    \item N-MNIST is a spiking version of the MNIST dataset \cite{53Orchard2015}. Same as the original dataset, it consists of 60,000 training samples and 10,000 testing samples corresponding to 10 classes.
\end{itemize}

\subsection{Network}
We evaluate the performance of the quantization method described earlier by building the SNN model formulated by \cite{51Fang2021}. This SNN model consists of the spiking encoder network and the classifier network, as shown in Fig. \ref{fig:Network}. A spiking encoder network is built from a convolutional layer, spiking neurons, and a pooling layer, whereas a classifier network - is from a fully connected dense layer and spiking neurons. Synaptic connections, represented by convolutional Conv2d and fully connected layers, are stateless, while spiking neuron layers have connections in the temporal domain, as depicted in  Fig. \ref{fig:Network}.  In pooling layers, max-pooling (MP) is utilized instead of commonly used average-pooling (AP). The output of the max-pooling layer is binary spikes, in contrast to floating-point numbers in average pooling, which, together with quantization, provide the possibility of accelerated computation.

\renewcommand{\arraystretch}{1.25}
\begin{table}[b!]
\centering
\caption{SNN network structure for each dataset.}
\label{tab:structure}
\begin{tabular}{ll}
\hline
Dataset        & Network                                                 \\ \hline
CIFAR10-DVS    & 5$\times$(128Conv3-MP2) - 512Dense100 - AP10  \\
DVS128 Gesture & 5$\times$(128Conv3-MP2) - 512Dense110 - AP10  \\
N-Caltech101   & 5$\times$(128Conv3-MP2) - 512Dense1010 - AP10 \\
N-MNIST        & 5$\times$(128Conv3-MP2) - 512Dense100 - AP10  \\ \hline
\end{tabular}
\end{table}

\begin{table}[b!]
\centering
\caption{SNN network structures adopted from \cite{48Eshraghian2022} for DVS128 Gesture and \cite{41Lui2021} for N-MNIST.}
\label{tab:structure_sota}
\begin{tabular}{ll}
\hline
Dataset        & Network                                                 \\ \hline
DVS128 Gesture & 16Conv5 - AP2 - 32Conv5 - AP2 - 8800Dense11  \\
N-MNIST        & 64Conv7 - MP2 - 128Conv7 - 128Conv7 - MP2 - Dense11   \\ \hline
\end{tabular}
\end{table}

\renewcommand{\arraystretch}{1.25}
\begin{table*}[t!]
\caption{Performance comparison between the proposed method and the state-of-the-art methods.}
\label{tab:SOTA}
\centering
\begin{tabular}{lllllll}
Dataset                       & Reference                                                 & Training Method                 & Quantization Method                   & Weights & Model Size (MB) & Accuracy (\%) \\ \hline 
\multirow{5}{*}{CIFAR10-DVS}  & \cite{54Wu2019}                          & STBP                            & -                                     & 32-bit  & 392             & 60.50         \\ \cline{2-7}
                              & \cite{55Wu2021}                          & Tandem learning                 & -                                     & 32-bit  & 416             & 58.65         \\ \cline{2-7}
                              & \cite{56Deng2020}                        & STBP                            & -                                     & 32-bit  & 53              & 60.30         \\ \cline{2-7}
                              & \multirow{2}{*}{\textbf{This work}}                                & \multirow{2}{*}{SGM} & \multirow{2}{*}{Linear Sigmoids}      & 32-bit  & 6.46            & \textbf{72.08}         \\  \cline{5-7}
                              &                                                           &                                 &                                       & 1-bit   & 0.21            & \textbf{64.05}         \\\hline \hline
\multirow{7}{*}{DVS128}       & \multirow{2}{*}{\cite{27Schaefer2020}}   & \multirow{2}{*}{DECOLLE}        & \multirow{2}{*}{Integer Quantization} & 32-bit  & -               & 88.89         \\  \cline{5-7}
                              &                                                           &                                 &                                       & 2-bit   & -               & 76.04         \\  \cline{2-7}
                              & \multirow{2}{*}{\cite{48Eshraghian2022}} & \multirow{2}{*}{Spike-based BP} & \multirow{2}{*}{Threshold annealing}  & 32-bit  & -               & 92.87         \\  \cline{5-7}
                              &                                                           &                                 &                                       & 1-bit   & -               & 91.32         \\  \cline{2-7}
                              &    \cite{49Amir2017}                        & CNN-to-SNN             & Deterministic rounding                & 2-bit   & 38              & 91.77         \\  \cline{2-7}
                              & \multirow{2}{*}{\textbf{This work}}                                & \multirow{2}{*}{SGM} & \multirow{2}{*}{Linear Sigmoids}      & 32-bit  & 6.48            & \textbf{96.63}         \\  \cline{5-7}
                              &                                                           &                                 &                                       & 1-bit   & 0.21            & \textbf{95.45}         \\\hline \hline
\multirow{4}{*}{N-Caltech101} & \cite{57she2022}                         & BPTT                            & -                                     & 32-bit  & 51.9             & 71.20         \\  \cline{2-7}
                              & \cite{58Kim2021}                         & SALT                            & -                                     & 32-bit  & -               & 55.00         \\  \cline{2-7}
                              & \multirow{2}{*}{\textbf{This work}}                                & \multirow{2}{*}{SGM} & \multirow{2}{*}{Linear Sigmoids}      & 32-bit  & 13.1            & \textbf{72.18}         \\ \cline{5-7}
                              &                                                           &                                 &                                       & 1-bit   & 0.41            & \textbf{68.71}         \\\hline  \hline
\multirow{7}{*}{N-MNIST}      & \multirow{2}{*}{\cite{41Lui2021}}        & \multirow{2}{*}{DECOLLE}        & \multirow{2}{*}{Stochastic rounding}  & 32-bit  & 4.84            & 98.10         \\   \cline{5-7}
                              &                                                           &                                 &                                       & 4-bit   & 0.61            & 81.10         \\  \cline{2-7}
                              & \cite{54Wu2019}                          & STBP                            & -                                     & 32-bit  & 392             & 99.53         \\  \cline{2-7}
                              & \cite{55Wu2021}                          & Tandem learning                 & -                                     & 32-bit  & 22              & 99.31         \\  \cline{2-7}
                              & \cite{56Deng2020}                        & STBP                            & -                                     & 32-bit  & 53              & 99.42         \\  \cline{2-7}
                              & \multirow{2}{*}{\textbf{This work}}                                & \multirow{2}{*}{SGM} & \multirow{2}{*}{Linear Sigmoids}      & 32-bit  &  4.84            & \textbf{99.63}        \\  \cline{5-7}
                              &                                                           &                                 &                                       & 1-bit   & 0.15            & \textbf{99.43}         \\ \hline 
\end{tabular}

\end{table*}

Two kinds of experiments are presented to analyze the performance of the presented quantization method. In the first one, a similar SNN model structure is trained on the four mentioned datasets. Five convolutional layers, each followed by a spiking neuron layer and max-pooling layer, are used for the spiking encoder network. A classifier network is built from two fully connected layers with spiking neurons, followed by a voting layer, which is a simple average-pooling layer with a window size of 10. Piecewise leaky ReLU is used as a surrogate function. The network structures for each dataset can be found in Table \ref{tab:structure}. In the second experiment, the performance of the quantization method is evaluated by using SNN models that are presented in other literature with SNN quantization method such as \cite{48Eshraghian2022} (DVS128 Gesture) and \cite{41Lui2021} (N-MNIST). These works are chosen for comparison because they achieve the highest quantized model accuracies. This selection minimizes the influence of a network structure and emphasizes the performance of the quantization method. Corresponding SNN model configurations can be found in Table \ref{tab:structure_sota}. In this work and \cite{48Eshraghian2022}, the SNN models are trained using SGM, while \cite{41Lui2021} uses DECOLLE \cite{42Kaiser2020}.

\subsection{Training}
The SNN models are trained on Nvidia V100 GPU for 500 epochs. An Adam optimizer with a learning rate of 0.001 is used during training. As a learning rate scheduler, we utilize a cosine annealing learning rate with the maximum number of iterations $T_{max}$ set to 64. The membrane time constant, $\tau$, is set to two for all spiking neurons. To evaluate the performance of the quantized SNN model, we choose four different bit-precision (fixed-point format) to compare with the full-precision: 8-bit, 4-bit, 2-bit (ternary), and 1-bit (binary). For the 8-bit precision, we use the quantization levels in the range \{-127, 127\}, where the gap between two levels is 1, e.g., \{-127, -126, …, -1, 0, 1, …, 126, 127\}. Similarly, for the 4-bit precision, the quantization levels lie in the range \{-7, 7\} (e.g., \{-7, -6, …, -1, 0, 1, …, 6, 7\}), for the ternary quantization there are three levels \{-1, 0, 1\}, and for the binary, there are two levels \{-1, 1\}. As can be seen, the quantization levels are chosen to be centered at 0 in all bit precision. The borders between two adjacent quantization levels are placed precisely in the middle. It is worth noting that these quantization levels and borders can also be placed non-uniformly. The temperature is initially set to T = 1, incrementing by 2 in each subsequent epoch. Quantized SNN models are evaluated using accuracy and memory savings metrics compared with the full-precision counterpart. Memory savings are defined by the compression ratio of the quantized models with respect to the full-precision models. Comparison with other state-of-the-art SNN models includes full and reduced bit precision works available in the literature.

\section{Results and Discussions}
\label{sec:Results}
\subsection{Comparison with prior works}
\label{subSOTA}
Table \ref{tab:SOTA} illustrates how the studied method compares with other full-precision and quantized SNN implementations. It shows the summary of learning rules and quantization methodology used in recent research works and highlights the bit precision, memory size, and final test accuracy.

\begin{figure*}[t!]
\begin{center}
\includegraphics[width=18cm]{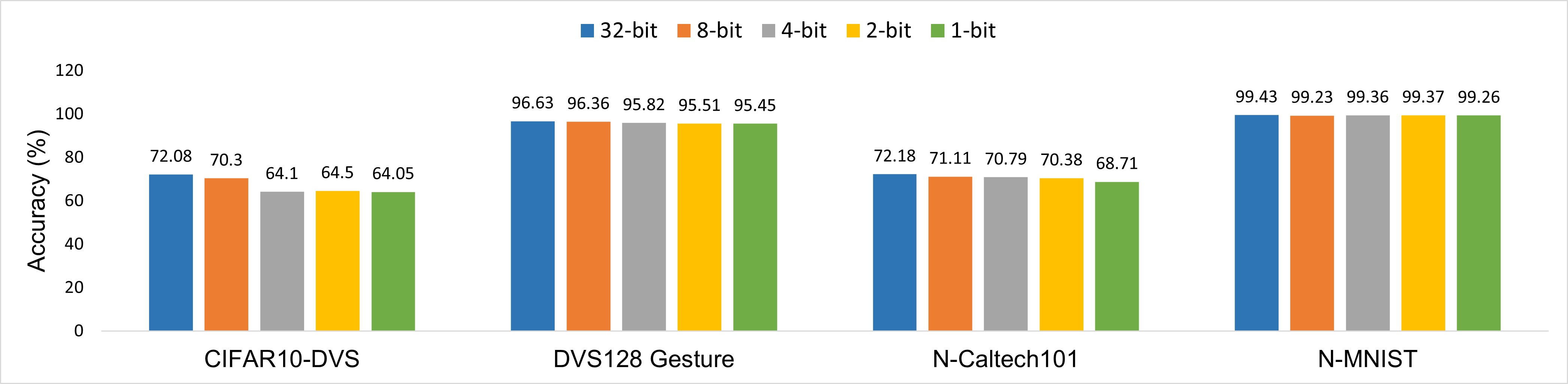}
\end{center}
\caption{Test accuracies of SNN models for each dataset.}\label{fig:figACCPrec}
\end{figure*}

\begin{figure*}[t]
\begin{center}
\includegraphics[width=18cm]{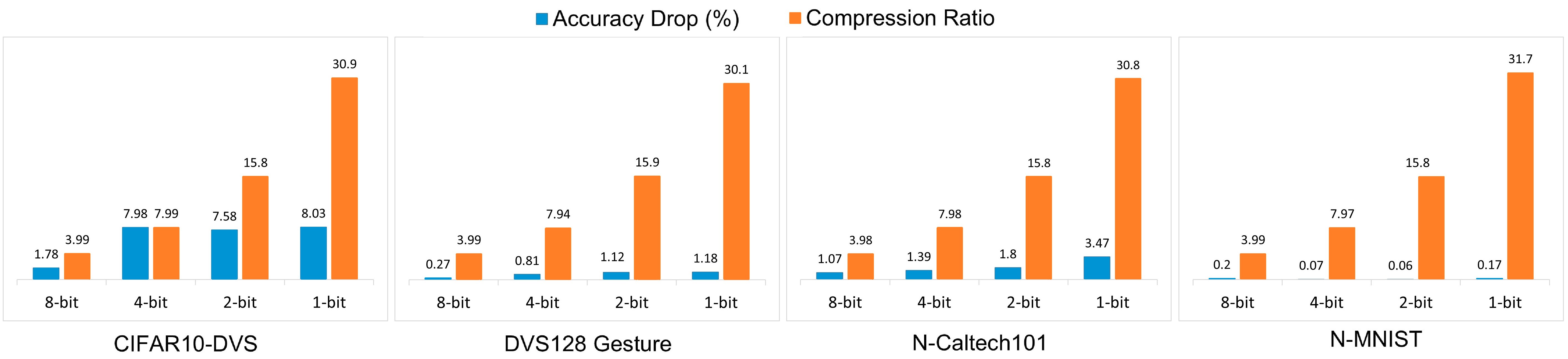}
\end{center}
\caption{Accuracy drop and compression ratio for quantized SNN models.} \label{fig:ACCMem}
\end{figure*}

\subsubsection{DVS128 Gesture}
For the DVS128 Gesture dataset, three quantization methodologies are available in the literature. Two of them reduce the bit precision to 2 bits, while the third binarizes the model. The full-precision model in this work shows higher accuracy than models reported in the literature. Also, our binarized model outperforms the 2-bit \cite{27Schaefer2020, 49Amir2017} and 1-bit\cite{48Eshraghian2022} cases shown in the table. 

Applying the proposed quantization method to the SNN model from \cite{48Eshraghian2022} allows for evaluating its performance based on accuracy. Table \ref{tab:DVS128a2a} presents accuracy results corresponding to 1-bit and 32-bit precisions. In this work, accuracies of both full-precision and 1-bit quantized SNN models are higher than the ones reported in \cite{48Eshraghian2022} by $1.44\%$ and $1.5\%$, respectively.

\begin{table}[]
\centering
\caption{Accuracy (\%) comparison between the proposed method and \cite{48Eshraghian2022} on DVS128 Gesture dataset with identical SNN model structure.}
\label{tab:DVS128a2a}
\begin{tabular}{|l|c|c|}
\hline
\multirow{2}{*}{Method} & \multicolumn{2}{c|}{Precision}                \\ \cline{2-3} 
 & {1-bit} & 32-bit \\ \hline
\cite{48Eshraghian2022}  & {91.32}      & 92.87  \\ \hline
This work         & {\textbf{92.82}} &  \textbf{94.31}  \\ \hline
\end{tabular}
\end{table}

\begin{table}
\centering
\caption{Accuracy (\%) comparison between the proposed method and \cite{41Lui2021} on N-MNIST dataset with identical SNN model structure.}
\label{tab:NMNISTa2a}
\begin{tabular}{|l|c|c|c|}
\hline
\multirow{2}{*}{Method} & \multicolumn{3}{c|}{Precision}                \\ \cline{2-4} 
& {1-bit} & {4-bit} & 32-bit \\ \hline
\cite{41Lui2021}& {-}     & {81.1}  & 98.1   \\ \hline
This work   & {\textbf{99.43}} & {\textbf{99.55}} & \textbf{99.63}  \\ \hline
\end{tabular}
\end{table}

\subsubsection{N-MNIST}
As seen in Table \ref{tab:SOTA}, our full-precision and binarized SNN model shows competitive performance compared to other full-precision and 4-bit quantized models. Only \cite{54Wu2019} with its 32-bit SNN model achieves slightly higher accuracy than our binarized model. 

The accuracy comparison between the proposed method and  \cite{41Lui2021} are shown in Table \ref{tab:NMNISTa2a}, where the SNN network structure is identical. Our binarized and 4-bit precision models outperform the 4-bit one from \cite{41Lui2021} by $18.33\%$ and $18.45\%$, respectively. At the same time, our full-precision model shows an accuracy that is $1.53\%$ higher than the one reported in \cite{41Lui2021}. It is worth mentioning that although the network structure is identical, the SNN training methods and selected optimization techniques are different in our work, which can explain the significant difference in the obtained results.

\begin{figure*}[t!]
\begin{center}
\includegraphics[width=16cm]{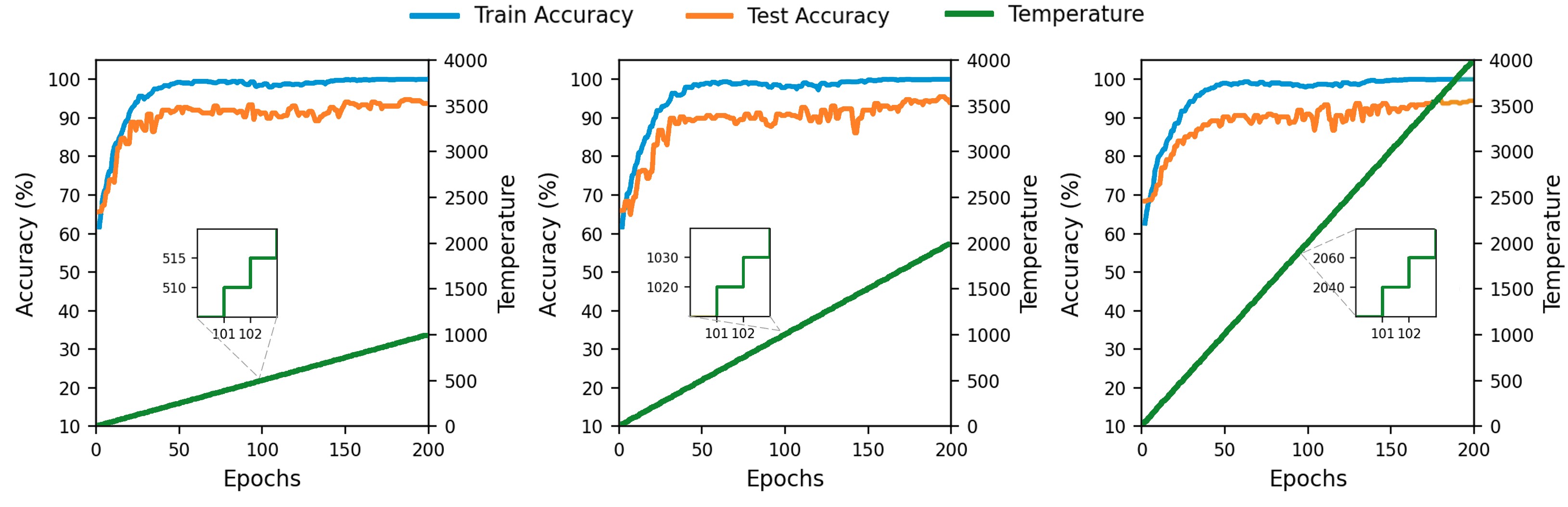}
\end{center}
\caption{Train and test accuracies of the binarized SNN model for different rates of temperature increase on the DVS128 Gesture dataset.} \label{fig:EffectofT_DVS}
\end{figure*}

\subsubsection{CIFAR10-DVS}
No quantized SNN model can be found in the literature for the CIFAR10-DVS dataset. Hence, the performance of the proposed quantized models is compared with other full-precision models. Both the 32-bit and 1-bit models in this work significantly outperform other works in terms of accuracy. Binarized model, while requiring almost 32 times less model size, still performs better than full-precision models with large model sizes, as reported in other publications. 

\subsubsection{N-Caltech101}
Since quantized models are not available in the case of the N-Caltech101 dataset, the models reported in this work are compared with other full-precision models in the literature. While the accuracy of our full-precision model is highest, our binarized model has slightly lower accuracy than the full-precision model in \cite{57she2022}. However, there is a substantial difference in memory size between our binarized model and other full-precision models while having comparable accuracy.

\subsection{Accuracy – bit-precision trade-off}
\label{subACCPrec}
Fig. \ref{fig:figACCPrec} illustrates the final test accuracy obtained for a particular combination of bit-precision value and dataset. The blue bar corresponds to the non-quantized, full-precision model and serves as a baseline for quantized models to be compared. Overall, the proposed quantization method provides similar performance for different classification tasks. Quantization has more impact on the SNN performance for more complex tasks, such as CIFAR10-DVS and N-Caltech101.

Quantization of the SNN model for the CIFAR10-DVS dataset yields the highest degradation in accuracy compared to other datasets. While the accuracy drop (1.78 \%) for the 8-bit model is not dramatic, performance degradation for lower-bit quantized models is more evident with accuracy drops of 7.98\%, 7.58\%, and 8.03\% for 4, 2, and 1-bit precisions respectively. The possible reason for such degradation can be the complex nature of the dataset.

The accuracy drop for the DVS128 Gesture dataset is much lower than for the CIFAR10-DVS dataset. The 8-bit quantized model achieves the lowest accuracy degradation (0.27\%). For the 4-bit quantization, the model experiences a 0.81\% drop in accuracy compared to the full precision model. For the 2-bit and 1-bit quantization, the accuracy drop is slightly higher than 1\% (1.12\% and 1.18\%, respectively). Hence, for this dataset, the model can be quantized down to 1-bit with an accuracy degradation of around 1\%.

For the N-Caltech101 dataset, on the other hand, the binarized SNN model has a more significant accuracy drop compared to other bit-precision models. While 8-bit, 4-bit, and 2-bit quantized SNN models have 1.07 \%, 1.39 \%, and 1.8 \% accuracy drops, the 1-bit model has 3.47 \% accuracy degradation compared to its full-precision counterpart. 

For the N-MNIST dataset, the accuracy degradation is the lowest among the presented datasets. The accuracy drop is negligible and fluctuates between different bit-precisions. The accuracy drop in the $< 0.2 \%$ range is achieved compared to the full-precision model.

\subsection{Accuracy drop – memory savings trade-off}
\label{subACCMemory}

The model size is calculated by multiplying the total number of parameters of the SNN model by the corresponding bit precision. Compression ratio (CR) is defined as the ratio of full-precision model size to quantization bit-precision model size. CR shows how much the model can be reduced when quantized to a particular bit-precision compared to the full-precision baseline. In Fig. \ref{fig:ACCMem}, amount of performance degradation can be observed to achieve a particular model size reduction. For the CIFAR10-DVS dataset, 8-bit quantization can reduce the model size by four times while keeping the accuracy degradation within 2\% from the baseline. In the DVS128 Gesture dataset, the most significant compression ratio, close to 31$\times$, is characterized by an accuracy drop slightly larger than 1\%. For the N-Caltech101 dataset, the SNN model can be reduced by 16$\times$ with the test accuracy degradation of 2\%. Finally, for the N-MNIST dataset, the SNN model shows the best accuracy-memory savings trade-off by having only a 0.2\% accuracy drop with a near 32-fold model size reduction.

\subsection{Effect of temperature rate on convergence}
\label{EffectOfT}
As discussed in Section \ref{sec:Method}, the temperature parameter $T$ controls the difference between sigmoid functions used during training Eq. (\ref{eq: training}) and step functions used in inference Eq. (\ref{eq: inference}). To investigate the effect of different temperature rates during training on convergence, we compare train and test accuracy with several values of $T$ for the binary precision, as it is the most challenging. As similar results are obtained for different datasets, we only include the results for one dataset. Fig. \ref{fig:EffectofT_DVS} shows the convergence of the binarized SNN model under three $T$ rates for the DVS128 Gesture dataset. Train and test accuracies are monitored for 200 epochs with $T = (5/10/20) \times epoch\, number$. The graphical representation of these increments is shown in the form of green lines. When the slope of these lines is greater, the temperature increase rate is higher. As it can be seen, the convergence of binary SNN models is stable w.r.t. to different rates of $T$; thus, it has a negligible effect on accuracy if given any reasonable value.

\section{Conclusion}
\label{sec:Conclusion}
In this work, we proposed an efficient method for the quantization of SNNs while focusing on the memory saving – accuracy drop trade-off. Using the differentiable quantization function, we can reduce the number of approximations for gradient computation when the SGM learning rule is employed. The obtained results demonstrate that using the proposed differentiable quantization function outperforms prior works based on rounding and threshold-adjusting techniques. The following results can be achieved using the linear combination of sigmoid functions as quantization function: binarized SNN models trained on CIFAR10-DVS, DVS128 Gesture, N-Caltech101, and N-MNIST show an accuracy drop of  8.03\%, 1.18\%, 3.47\%, and 0.17\%, respectively, (as compared with full-precision counterparts) while providing up to 31$\times$ memory savings.

\section{Acknowledgment}
\label{sec:Ack}
This work was supported by The King Abdullah University of Science and Technology under the award ORA-2021-CRG10-4704.


\end{document}